# Can Uncertainty Management Be Realized In A Finite Totally Ordered Probability Algebra?

Yang Xiang*, Michael P. Beddoes* and David Poole[†]
University of British Columbia, Vancouver, B.C., Canada, V6T 1W5
* Department of Electrical Engineering, yangx@sun280.ee.ubc.ca
[†] Department of Computer Science, poole@cs.ubc.ca

## Abstract

In this paper, the feasibility of using finite totally ordered probability models under Aleliunas's Theory of Probabilistic Logic [Aleliunas 88] is investigated. The general form of the probability algebra of these models is derived and the number of possible algebras with given size is deduced. Based on this analysis, we discuss problems of denominator-indifference and ambiguity-generation that arise in reasoning by cases and abductive reasoning. An example is given that illustrates how these problems arise. The investigation shows that a finite probability model may be of very limited usage.

## 1 Introduction

This research started from the process of building a medical diagnostic expert system, in the domain of EEG analysis. In this domain we wanted to combine evidence, but the experts consulted claimed that they did not use numbers, but rather used a small number of terms to describe uncertainty. Thus we were lead to a finite non-numerical uncertainty management mechanism. In such a mechanism, the domain expert's vocabulary about uncertainty could be used directly in encoding knowledge and in reasoning about uncertain information. This would facilitate knowledge acquisition and make the system's diagnostic suggestion and explanation more understandable.

There were few known finite mechanisms for general uncertainty management [Halpern 87, Pearl 89], but we were drawn to Aleliunas' probabilistic logic [Aleliunas 88], because it seemed to be based on clear intuitions where measures of belief (probability values) could be summarised by values other than just real numbers.

Aleliunas [Aleliunas 88] presents an axiomatization for a theory of rational belief, the *Theory of Probabilistic Logic (TPL)*. It generalizes classical probability theory to accommodate a variety of probability values rather than just $[0,1]$. According to the theory, *probabilistic logic* is a scheme for relating a body of evidence to a potential conclusion (a hypothesis) in a rational way, using *probabilities* as *degrees of belief*. '$p(P|Q)$' stands for the conditional probability of proposition $P$ given the evidence $Q$, where $P$ and $Q$ are sentences of some formal language $L$ consisting of boolean combinations of propositions. TPL is chiefly concerned with identifying the characteristics of a family of functions from $L \times L$ to the set of probabilities $\mathbf{P}$. The probability values $\mathbf{P}$ are not constrained to be just $[0,1]$, but can be any values that conform to a set of reasonably intuitive axioms [Aleliunas 88].

The semantics of TPL is given by 'possible worlds'. Each proposition $P$ is associated with a set of *situations* or *possible worlds* $S(P)$ in which $P$ holds. Given $Q$ as evidence, the conditional probability $p(P|Q)$, whose value ranges over the set $\mathbf{P}$, is some measure of the fraction of the set $S(Q)$ that is occupied by the subset $S(P\&Q)$.

TPL provided minimum constraints for a rational belief model. For our particular domain we thought the following criteria were desirable:

**R1** The domain experts did not believe that they used numerical values for uncertainty. Their language consisted of a small set of terms "likely", "possibly", etc., used to describe the uncertainty in their domain. Thus we were lead to a **finite** set of probability values.

**R2** Any two probability values in a chosen model should be comparable. An essential task of a medical diagnostic system is to estimate the likelihood of a set of competing diagnoses given a patient's symptoms and history. We felt as though we needed to have totally ordered probabilities in order to allow for totally ordered decisions when we have to act on the results of the diagnoses.

**R3** Inference based on a TPL model should generate empirically sound results. That is, the inference outcomes generated with such a model should reflect, as far as possible, the reasonable outcomes reached by a human expert.

**R4** We require some reasonable statistical assump-tions. For our domain, the assumptions embedded in Bayesian networks [Pearl 88] seemed to be particularly appealing. Although this statistical assumption was used in our implementation, the analysis presented here does not seem to critically depend on the statistical assumptions used.

Although these criteria are formed from the point of our application, we believe they are shared by many automated reasoning systems making decisions under uncertainty.

Based on these criteria, we concentrate on finite totally ordered probability models.

## 2 Finite totally ordered probability algebras

### 2.1 The algebra of probability values

To investigate the mathematical structure (*probability algebra*) of the probability space, the characterization of any finite totally ordered probability algebra under TPL axioms [Aleliunas 88] is given, without proof, in the proposition below (for more about universal algebra, see [Burris 81, Kuczkowski 77]). This proposition is a restriction of a general theorem in Aleliunas [Aleliunas 86] to finite totally ordered sets.

We denote the smallest element of $P$ by 0, and the largest element of $P$ as 1. There are a finite number of other values between 0 and 1.

**Proposition 1** *A probability algebra defined on a totally ordered finite set $P$ satisfies TPL axioms iff*

1. *An order preserving binary operation "$*$" (product) is well defined and closed on $P$.*
2. *"$*$" is commutative, i.e. $(\forall p, q \in P)\ p * q = q * p$.*
3. *"$*$" is associative, i.e. $(\forall p, q, r \in P)\ p * (q * r) = (p * q) * r$.*
4. *$(\forall p, q, r \in P)\ (p * q = r) \Rightarrow (r \leq \min(p, q))$.*
5. *No non-trivial zero, i.e. $(\forall p, q \in P)\ p * q = 0 \Rightarrow (p = 0 \lor q = 0)$.*
6. *$(\forall p, q \in P)\ p \leq q \Rightarrow (\exists r \in P)\ p = r * q$. The solution will be denoted as $r = p/q$.*
7. *$(\forall p \in P)\ 0 \leq p \leq 1$.*
8. *$(\forall p \in P)\ p * 1 = p$.*
9. *A monotone decreasing inverse function $i[\cdot]$ is well defined and closed on $P$, i.e. $(\forall p < q \in P)\ i[p] > i[q]$.*
10. *$(\forall p \in P)\ i[i[p]] = p$.*

From now on any *Finite Totally Ordered Probability Algebra* satisfying proposition 1 will be referred as *legal FTOPA*. The general form of all legal FTOPA will be derived in next section.

### 2.2 Mathematical structure

Here we are interested in only those probability algebras with at least 3 elements[1]. A finite totally ordered probability set with size $n$ is denoted as $P = \{e_1, e_2, \ldots, e_{n-1}, e_n\}$, where $1 = e_1 > e_2 > \ldots > e_{n-1} > e_n = 0$. For example, $P = \{e_1, e_2, e_3, e_4\}$ could stand for {certain, likely, unlikely, impossible}. This linguistic interpretation is left open.

The uniqueness of the inverse function $i[\cdot]$ of any legal FTOPA is given by the following lemma.

**Lemma 1** *For a legal FTOPA with size $n$, the inverse is uniquely defined as*

$$i[e_k] = e_{n+1-k} \quad (1 \leq k \leq n).$$

Thus given the size of a legal FTOPA, only the choice of the product function is left.

A probability $p \in P$ is *idempotent* if $p * p = p$. idempotent elements play important roles in defining probability algebras as will be shown in a moment.

**Lemma 2** *Any legal FTOPA has at least 3 idempotent elements, namely $e_1$, $e_{n-1}$ and $e_n$.*

This lemma is required in the proof of proposition 2 and corollary 1.

**Lemma 3** *For any legal FTOPA, if $p \in P$ is idempotent, then $(\forall q \in P)\ p * q = \min(p, q)$.*

Lemma 3 is important in the proof of proposition 2, and theorem 1. Aleliunas[Aleliunas 86] gives similar statement.

**Proposition 2** *For a finite totally ordered set with size $n \geq 3$, there exists only one legal FTOPA with 3 idempotent elements. The "$*$" operation on it is defined as*

$$e_i * e_j = \begin{cases} e_n & \text{if } i \text{ or } j = n \\ e_{\min(i+j-1, n-1)} & \text{otherwise.} \end{cases}$$

Proof:

Let $M_{n,k}$ denote a legal FTOPA with size n and k idempotent elements.[2] Let $a_{i,j}$ denote $e_i * e_j$. We prove the proposition constructively.

(1) In case of $i$ or $j = n$, the proposition holds due to lemma 3. By non-trivial zero, zero part of the product table is entirely covered within this case.

(2) What is left is to prove the non-zero part of the product table (the second half of the product formula) which is bounded by two idempotent elements $e_1$ and $e_{n-1}$. For the completeness of the product table, we still include the zero parts in the following tables although they are not relevant to the remaining proof.

---

[1] Probability algebra with 2 elements is equivalent to propositional logic [Aleliunas 87].

[2] In general, for a pair of n and k, there may be more than one legal FTOPA. Thus $M_{n,k}$ does no necessarily stand for a unique model characterized by n and k.

For $M_{3,3}$ and $M_{4,3}$ the proposition holds (see the product tables below). It is not difficult to check that they satisfy proposition 1 and any change to these product tables will violate proposition 1 in one way or another.

|       | $e_1$ | $e_2$ | $e_3$ |
|-------|-------|-------|-------|
| $e_1$ | $e_1$ | $e_2$ | $e_3$ |
| $e_2$ | $e_2$ | $e_2$ | $e_3$ |
| $e_3$ | $e_3$ | $e_3$ | $e_3$ |

$M_{3,3}$

|       | $e_1$ | $e_2$ | $e_3$ | $e_4$ |
|-------|-------|-------|-------|-------|
| $e_1$ | $e_1$ | $e_2$ | $e_3$ | $e_4$ |
| $e_2$ | $e_2$ | $e_3$ | $e_3$ | $e_4$ |
| $e_3$ | $e_3$ | $e_3$ | $e_3$ | $e_4$ |
| $e_4$ | $e_4$ | $e_4$ | $e_4$ | $e_4$ |

$M_{4,3}$

Suppose a unique legal FTOPA $M_{m,3}$ exists with product defined as in the proposition. As for $M_{m+1,3}$ (table below), the product $a_{i,j}$ $(i+j \leq m)$ should be constructed in the same way as in $M_{m,3}$, i.e. the second half of product formula

$$a_{i,j} = e_{\min(i+j-1,m)} = e_{i+j-1}$$

applies within this portion as does in $M_{m,3}$. If this portion could be changed without violating proposition 1, the corresponding portion in $M_{m,3}$ could also be changed which is contradictory to the uniqueness assumption for $M_{m,3}$.

Further we show the uniqueness of $a_{i,j}$ for all $(i,j < m < i+j)$.

|           | $e_1$ | $e_2$ | $e_3$ | ... |         |             | $e_{m-1}$ | $e_m$ | $e_{m+1}$ |
|-----------|-------|-------|-------|-----|---------|-------------|-----------|-------|-----------|
| $e_1$     | $e_1$ | $e_2$ | $e_3$ | ... |         |             | $e_{m-1}$ | $e_m$ | $e_{m+1}$ |
| $e_2$     | $e_2$ | $e_3$ | ...   |     |         | $e_{m-1}$   | $a_{2,m-1}$ | $e_m$ | $e_{m+1}$ |
| $e_3$     | $e_3$ | ...   |       |     | $e_{m-1}$ | $a_{3,m-2}$ | $a_{3,m-1}$ | $e_m$ | $e_{m+1}$ |
| ⋮         | ⋮     | ...   | $a_{j,m-j+1}$ $a_{j+1,m-j}$ | | | | | ⋮ | ⋮ |
| $e_{m-1}$ | $e_{m-1}$ | ... | | | | ? | | | |
| $e_m$     | $e_m$ | | | ... | | | | | $e_m$ |
| $e_{m+1}$ | $e_{m+1}$ | | | ... | | | | | $e_{m+1}$ |

Note: "?" stands for product items to be chosen.

$M_{m+1,3}$

By associativity, we have

$$\begin{aligned}
(e_j * e_2) * e_{m-j} &= e_{j+1} * e_{m-j} \\
&= a_{j+1,m-j} \\
&= e_j * (e_2 * e_{m-j}) = e_j * e_{m-j+1} \\
&= a_{j,m-j+1} \qquad (2 \leq j \leq m-2) \quad (a)
\end{aligned}$$

Also we have

$$\begin{aligned}
e_2 * (e_j * e_{m-1}) &= e_2 * a_{j,m-1} \\
&= (e_2 * e_j) * e_{m-1} = e_{j+1} * e_{m-1} \\
&= a_{j+1,m-1} \qquad (2 \leq j \leq m-2) \quad (b)
\end{aligned}$$

From order preserving property of "$*$", we know

$$a_{i,j} = e_{m-1} \quad \lor \quad a_{i,j} = e_m \quad (i,j < m < i+j).$$

Suppose $a_{2,m-1} = e_{m-1}$. Then from (b),

$$e_2 * a_{2,m-1} = e_2 * e_{m-1} = a_{2,m-1} = e_{m-1} = a_{3,m-1}.$$

Similarly, and from commutativity and order preserving, we have

$$a_{i,j} = e_{m-1} \quad (i,j < m < i+j).$$

This means that $e_{m-1}$ is also an idempotent element which is contradictory to the 3 idempotent elements assumption. Therefore, $a_{2,m-1} = e_m$. Then from (a) and order preserving, we end up with $a_{i,j} = e_m$ $(i,j < m < i+j)$.

□

The second part of the above proof for product bounded by $e_1$ and $e_{n-1}$ does not involve the 0 element at all as already stated. Thus for any legal FTOPA with more than 3 idempotent elements, the proposition holds for each diagonal block of its product table bounded by two adjacent idempotent elements. The non-diagonal part of the product table is totally determined by lemma 3, the order preserving and solution existing property. Thus we have the following theorem 1. Given proposition 2 and above description, the proof is trivial.

**Theorem 1** *Given a finite totally ordered set $P = \{e_1, e_2, \ldots, e_n\}$ with ordering relation $e_1 > e_2 > \ldots > e_n$ and a set $I$ of indexes of all the idempotent elements on $P$, $I = \{i_1, i_2, \ldots, i_m\}$ where $i_1 < i_2 < \ldots < i_m$ there exists a unique legal FTOPA whose product function is defined as:*

$$e_j * e_k = \begin{cases} \min(e_j, e_k) & \text{if } e_j = i_l \\ e_{\min(j+k-i_l, i_{l+1})} & \text{if } i_l < j, k \leq i_{l+1} \\ e_k & \text{if } j \leq i_l < k \leq i_{l+1} \end{cases}$$

*and inverse function is defined as:*

$$i[e_k] = e_{n+1-k}$$

Theorem 1 says that, given the set of idempotent elements, a legal FTOPA is totally defined. From theorem 1 and lemma 2 we can easily derive the following corollary.

**Corollary 1** *The number of all the possible legal FTOPA of size $n \geq 3$ is*

$$\sum_{i=0}^{n-3} C_{n-3}^i = 2^{n-3}$$

*where $C_m^i$ is the number of combinations taking $i$ elements out of $m$.*

Theorem 1 and corollary 1 provide the possibility of exhaustive investigation for any legal FTOPA of a given size.

## 2.3 Solution and range

Once a legal FTOPA is defined, its solution table is forced. Inverses to the operation $* : P \times P \to P$ will not be unique. For this reason, it is necessary to introduce a probability *range* denoted by $[l, u]$ representing all the probability values between lower bound $l$ and upper bound $u$.

$$[l, u] = \{v \in P | l \leq v \leq u\}$$

We write $[v, v]$ as just $v$. The following corollary on single value probability solution is given without proof.

**Corollary 2** *Given a finite totally ordered set $P = \{e_1, e_2, \ldots, e_n\}$ with ordering relation $e_1 > e_2 > \ldots > e_n$ and a set $I$ of indexes of all the idempotent elements on $P$, $I = \{i_1, i_2, \ldots, i_m\}$ where $i_1 < i_2 < \ldots < i_m$ the solution function (multiple value) of a legal FTOPA is forced to be:*

$e_k/e_j =$

$$\begin{cases} e_k & \text{if } j = 1 \\ e_n & \text{if } k = n, j \neq n \\ e_{k-j+i_l} & \text{if } k > j, \\ & \quad i_l + 1 < k \leq i_{l+1} - 1, \\ & \quad i_l + 1 \leq j < i_{l+1} - 1 \\ [e_{i_l}, e_1] & \text{if } k = j, i_l + 1 \leq k < i_{l+1} \\ [e_{i_{l+1}}, e_{i_{l+1}+i_l-j}] & \text{if } k = i_{l+1}, i_l < j < i_{l+1} \\ [e_k, e_1] & \text{if } k = j = i_l, k \neq 1, n \\ e_k & \text{if } j \leq i_l < k \leq i_{l+1} \end{cases}$$

In Appendix B, the product and solution tables for 3 legal FTOPAs of size 8 are presented.

The solution of two single valued probabilities may become a range which will participate in further manipulation. Thus the product and solution of ranges should be considered before we can manipulate uncertainty in an inference chain.

**Definition 1** *For any legal FTOPA, the product of two ranges $[a, b]$ ($a \leq b$) and $[c, d]$ ($c \leq d$) is defined as*

$$[a, b] * [c, d] = \{z | \exists x \in [a, b] \& \exists y \in [c, d] \& z = x * y\}.$$

*And the solution of above two ranges with additional constraint $a \leq d$ is defined as*

$$[a, b]/[c, d] = \{z | \exists x \in [a, b] \& \exists y \in [c, d] \& x = y * z\}.$$

One can prove the following proposition:

**Proposition 3** *For any legal FTOPA, the product of two ranges $[a, b]$ ($a \leq b$) and $[c, d]$ ($c \leq d$) is*

$$[a, b] * [c, d] = [a * c, b * d].$$

*And the solution of above two ranges with additional constraint $a \leq d$ is*

$$[a, b]/[c, d] = \begin{cases} [LB(a/d), UB(b/c)] & \text{if } b \leq c \\ [LB(a/d), e_1] & \text{if } b > c \end{cases}$$

*where $LB$ and $UB$ are lower and upper bounds of ranges.*

It should be noted that, in general, product and solution of legal FTOPAs do not follow commutativity. For example, in model $M_{8,8}$,

$$(e_2 * e_5)/e_5 = [e_5, e_1] \neq e_2 * (e_5/e_5) = [e_5, e_2].$$

Thus the order of product and solution in evaluation of conditional probability

$$p(A|B\&C) = p(A\&B|C)/p(B|C)$$
$$= (p(B|A\&C) * p(A|C))/p(B|C)$$

can not be changed arbitrarily.

## 3 Bayes theorem and reasoning by case

Having derived the mathematical structure of legal finite totally ordered probability models, we need deductive rules. In this investigation, we adopted Bayesian Networks [Pearl 88] (using the implementation described in [Poole 88]) as our scheme of knowledge representation. The inferencing rules required within this scheme are *Bayes theorem* and *reasoning by cases*.

Bayes theorem provides a way of determining the likelihood of certain causes from the observation of effects.[3] It takes the form:

$$p(P|Q\&C) = p(Q|P\&C) * p(P|C)/p(Q|C)$$

Reasoning by cases is an inference rule to compute a conditional probability by partitioning the condition into several exclusive situations such that the estimation under each of them is more manageable. The simplest form is considering the cases where $B$ is true and where $B$ is false:

$$p(A|C) = p((A\&B) \vee (A\&\overline{B})|C)$$

Under classical probability theory, it becomes:

$$p(A|C) = p(A|B\&C) \cdot p(B|C) + p(A|\overline{B}\&C) \cdot p(\overline{B}|C)$$

Using TPL, the $\cdot$ becomes $*$, and we do not have the $+$. This can, however, be simulated using product and inverse. The corresponding formula under TPL is given by the following propositions.

**Proposition 4** *Let $A$, $B$, and $C$ be three sentences. $p(A|C)$ can be computed using the following:*

$$\begin{aligned} f_1 &= i[p(A|\overline{B}\&C) * i[p(B|C)]] \\ f_2 &= i[p(A|B\&C) * p(B|C)/f_1] \\ p(A|C) &= i[f_1 * f_2]. \end{aligned}$$

Operationally, the computation of the likelihood of a hypothesis given some set of evidence using Bayesian Networks and legal FTOPAs is to apply two inference rules, namely, Bayes theorem and reasoning by cases [Pearl 88][Poole 88].

---

[3] *Cause* and *effect* are used here in a very weak sense.

# 4 Problems with legal finite totally ordered probability

## 4.1 Ambiguity-generation and denominator-indifference

Now that we have derived the mathematical structure of legal finite totally ordered probability models and the form of relevant deductive rules, we can assess these probability models as to how well they fit in with our intuition.

To begin with, we examine the solution of legal FTOPA $M_{n,n}$ which has all its elements idempotent. The solution takes the form of (compare to Appendix B)

$$e_k/e_j = \begin{cases} e_k & \text{if } j < k \\ [e_k, e_1] & \text{if } k = j \end{cases}$$

Note that $e_j$ does not have direct influence on the result of the first case of the solution. We name this phenomenon as *denominator-indifference*. Also, we name the emergence of range in the second case of the solution operation as *ambiguity-generation*.

To analyze the effect of denominator-indifference and ambiguity-generation on application of Bayes theorem, apply Bayes theorem to $M_{n,n}$.

$p(A|B\&C)$
$= p(A\&B|C)/p(B|C)$
$= \begin{cases} p(A\&B|C) & \text{if } p(A\&B|C) \neq p(B|C) \\ [p(A\&B|C), e_1] & \text{if } p(A\&B|C) = p(B|C) \end{cases}$

In the first case, the prior $p(B|C)$ does not affect the estimation of $p(A|B\&C)$ due to denominator-indifference. In the second case, ambiguity-generation produces a disjunct of all the probabilities larger than $p(B|C)$ which is a very rough estimation. Neither satisfies our requirement for empirically satisfactory probability estimates.

To analyze the effect of denominator-indifference and ambiguity-generation on reasoning by cases, consider applying proposition 4 to $M_{n,n}$.

$p(A|C)$
$= \begin{cases} \max(p(A\&B|C), p(A\&\overline{B}|C)) \\ \quad \text{if } p(A\&B|C) \neq p(A\&\overline{B}|C) \\ [\max(p(A\&B|C), p(A\&\overline{B}|C)), e_1] \\ \quad \text{if } p(A\&B|C) = p(A\&\overline{B}|C) \end{cases}$

Here again, in the first situation, denominator-indifference forces a choice of outcome from one case or another instead of giving some combination of the two outcomes. We do not get an estimation larger than both which is contrary to our intuition. In the second situation, a very rough estimation appears because of ambiguity-generation. Note that, when $\max(p(A\&B|C), p(A\&\overline{B}|C))$ is small, $p(A|C)$ can span almost the whole range of probability set **P**.

The analysis here was in terms of a model that has all of its values idempotent. The other case to consider is what happens at the values between the idempotent values.

Consider $M_{n,3}$ which has minimal number of idempotent elements. By proposition 2, its product is

$$e_i * e_j = \begin{cases} e_{\min(i+j-1,n-1)} & \text{if } i,j \neq n \\ e_n & \text{otherwise.} \end{cases}$$

Its solution simplifies to (compare to Appendix B)

$$e_k/e_j = \begin{cases} e_n & \text{if } k = n > j \\ e_{k-j+1} & \text{if } j \leq k < n-1 \\ [e_{n-1}, e_{n-j}] & \text{if } k = n-1 \geq j \end{cases}$$

In this algebra, it is quite easy for a manipulation to reach the probability value $e_{n-1}$:

1. Whenever one of the factors of product is $e_{n-1}$, the product will be $e_{n-1}$ unless the other factor is $e_n$.

2. Whatever takes the value $e_2$, its inverse will be $e_{n-1}$.

3. Products of low or moderate probability tend to reach $e_{n-1}$ due to quick decreasing of product.

4. $e_j/e_{j-1} = e_2$ for all $2 \leq j \leq n-2$.

Once $e_{n-1}$ is reached, any solution will be ambiguous. This ambiguity will be propagated and amplified during further inference in Bayesian analysis or case analysis. Although $e_{n-1}$ is a value we should try to avoid, we have no means to avoid it. Here we see an interesting trade off between the two problems. In $M_{n,3}$, denominator-indifference disappears. But, since manipulations under this model move probability values quickly, we tend to produce $e_{n-1}$ more frequently and thus suffer more from ambiguity-generation.

As all finite totally ordered probability algebras can be seen as combinations of the above two cases, they must all suffer from denominator-indifference and ambiguity-generation. The question now is how serious are the problems in an arbitrary model? This is to be answered in next section.

## 4.2 Quantitative analysis of the problems

Given the constraint of legal FTOPA in choosing a probability model, we are free to select the model size n and to select among $2^{n-3}$ alternative legal FTOPAs once n is fixed. We introduce a few straightforward measurements to quantify the degree of suffering in a randomly chosen model.

The number of ranges in a model's solution table and the number of elements covered by each range mirror the problem of ambiguity-generation of the model. Thus we define a measurement of the amount of ambiguity in a model as the number of elements covered by ranges in its solution table minus the number of ranges.

**Definition 2** *Let $S = \{r_1, r_2, \ldots, r_m\}$ be the set of ranges in the solution table of a legal FTOPA. Let $w_j$ be the number of values covered by range $r_j$. Let $M$ be the number of different solution pairs in the solution table.*

*The* **amount of ambiguity** *of the algebra is defined as*

$$A = \sum_{j=1}^{m} w_j - 1.$$

*The* **relative ambiguity** *of the algebra is defined as*

$$R = A/M.$$

We have the following proposition:

**Proposition 5** *The amount of ambiguity of any legal FTOPA with size n is*

$$A = (n-1)(n-2)/2.$$

*The relative ambiguity of the algebra is*

$$R = (n-2)/(n+2).$$

The number of solution pairs satisfying $e_j/e_k = e_j$ reflects the seriousness of denominator-indifference of the model. We define the order of denominator-indifference as this number minus the number of such $e_j$s.

**Definition 3** *Let $d_j$ be the number of times $e_j/e_k = e_j$ for $1 \leq k \leq j$ in a legal FTOPA of size n. The* **order of denominator-indifference** *of the algebra is defined as*

$$O_d = \sum_{j=2}^{n-1} d_j - 1.$$

We also define the order of mobility of a model to express the likelihood of a product or a solution transferring operands to different value. The higher this order, the more likely for a manipulation to generate an idempotent element and produce ambiguity afterwards.

**Definition 4** *The* **order of mobility** $O_m$ *of a legal FTOPA is defined as the number of distinct product pairs $a*b$ in its product table such that $a*b < \min[a, b]$.*

We have the following proposition:

**Proposition 6** *For any legal FTOPA with size n and a set I of indexes of all its idempotent elements $I = \{i_1, i_2, \ldots, i_k\}$ where $i_1 < i_2 < \ldots < i_k$ its order of denominator-indifference is*

$$O_d = \sum_{m=2}^{k-2} (i_m - 1) \cdot (i_{m+1} - i_m),$$

*its order of mobility is*

$$O_m = \sum_{m=1}^{k-2} \sum_{j=1}^{i_{m+1}-i_m-1} j,$$

*and*

$$O_d + O_m = (n-2)(n-3)/2.$$

Proposition 5 tells us that all the legal FTOPAs of same size have same amount of ambiguity. Increasing size *increases* $R$ which approaches 1 as $n$ gets larger and larger.

Proposition 6 says that,

1. among legal FTOPAs of same size n, the order of denominator-indifference $O_d$ changes from lower bound 0 at $M_{n,3}$ to upper bound $(n-2)(n-3)/2$ at $M_{n,n}$;

2. the upper bound of $O_d$ as well as $O_m$ increases with model size n;

3. given n, the sum $O_d + O_m$ remains constant and thus if a model suffers less from denominator-indifference, it must suffer more frequently from ambiguity-generation due to the increase in its mobility.

### 4.3 Can the changes in priors help ?

After we have explored model size and alternative models given size, the final freedom that remains is the assignment of prior probability values. From Corollary 2, it is apparent that, in general, denominator-indifference and ambiguity-generation happen only in certain regions of the solution table. So, is it possible, by choosing certain set of probability values as prior knowledge, to avoid intermediate results falling onto those unfavorable regions?

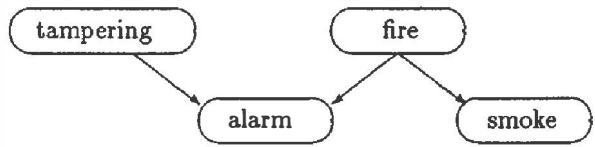

Fig. 1 Smoke-alarm example

To help answer this question, a derivation of conditional probability $p(fire|smoke\&alarm)$ for a smoke-alarm problem[4] in Fig. 1 is given in Appendix A. The calculation involves 2 applications of Bayes theorem, and 3 of reasoning by cases. It requires 19 products, 9 solutions, and 14 inverses.

In general,

1. a product tends to decrease the probability value until an idempotent value is reached.

2. a solution tends to increase the probability value or cause a large range to occur (especially for idempotent values).

---

[4]This four nodes example is the minimum one which has alternative hypotheses (fire and tampering) and allow accumulation of evidences (smoke + alarm).

3. an inverse tends to transfer small value into big and vice versa.

Since many operations are required even in a small problem and each operation tends to move the intermediate value around the probability set, the compound effect of the operations are not generally controllable.

To summarize, in the context of legal FTOPA, there seems to be no way to get away with the problem of denominator-indifference and ambiguity-generation by means of clever assignment of prior probability values; increasing model size does no good in reducing the difficulty; selecting among different models trades one trouble with another.

In the next section, these problems are demonstrated by an experiment.

## 5 An experiment

All the 32 legal FTOPAs with size 8 were implemented in Prolog and their performance are tested by the smoke-alarm example in [Poole 88] [5] The Prolog program has basically the same structure as [Poole 88], but inverse, product, solution, as well as Bayes theorem and reasoning by cases are redefined.

The following are part of the priors and conditional probabilities in the knowledge base together with numerical values used in [Poole 88] as a comparison.

| | | |
|---|---|---|
| $p(fire) =$ | $e_6$ | 0.01 |
| $p(tampering) =$ | $e_6$ | 0.02 |
| $p(smoke\|fire) =$ | $e_2$ | 0.9 |
| $p(smoke\|\overline{fire}) =$ | $e_6$ | 0.01 |
| $p(alarm\|fire\&tampering) =$ | $e_4$ | 0.5 |
| $p(alarm\|fire\&\overline{tampering}) =$ | $e_2$ | 0.99 |
| $p(alarm\|\overline{fire}\&tampering) =$ | $e_2$ | 0.85 |
| $p(alarm\|\overline{fire}\&\overline{tampering}) =$ | $e_7$ | 0.0001 |

*Table* 1

The following are some of the conditional probabilities calculated in all 32 possible legal FTOPAs with size 8 and in [0, 1] real number probability model as a comparison.

$$p(s|f), p(a|f), p(s|t), p(a|t),$$

$$p(f|s), p(f|a), p(f|s\&a), p(t|s), p(t|a), p(t|s\&a)$$

The first 4 probabilities are deductive which, given cause acting, estimate the likelihood of effects appearing. The remaining 6 are abductive which, given effects observed, estimate the likelihood of each conceivable cause.

- Among the 32 legal FTOPAs, 8 of them produced identical value for the abductive cases:

  $$p(f|s) = p(f|a) = p(f|s\&a) = e_6,$$

[5] Example in Fig. 1 is a subproblem of it.

and 16 others produce the identical ranges for all the abductive cases about fire.

From the knowledge in Table 1, we know that smoke does not necessarily relate to fire ($p(s|\overline{f}) = e_6$). Nor does alarm ($p(a|\overline{f}\&t) = e_2$). As a result, observing only one of smoke and alarm, we are not quite sure about fire. Intuitively, adding the positive evidence alarm to smoke should increase our belief for fire. As well, adding to alarm the evidence smoke which is independent of tampering indicates higher likelihood of fire causing alarm. Thus this intuitive inference arrives at

$$p(f|s\&a) > p(f|s) \quad \& \quad p(f|s\&a) > p(f|a)$$

which the results obtained from the above mentioned 24 legal FTOPAs do not fit in with.

To illustrate how this happens, evaluate $p(f|s\&a)$ in model $M_{8,4}$ with idempotent elements $\{e_1, e_5, e_7, e_8\}$.

$$p(f|s\&a) = p(s|f\&a) * p(f|a)/p(s|a)$$
$$= e_2 * e_6/e_4 = e_6/e_4 = e_6$$

Pay attention to the solution in last step. The result is no larger than $p(f|a) = e_6$ due to denominator-indifference. We do not get extra evidence accumulating.

- One of the very useful results provided by [0, 1] numerical probability is that although $p(f|s) = 0.48$ and $p(f|a) = 0.37$ are moderate, when both smoke and alarm are observed $p(f|s\&a) = 0.98$ is quite high which is more intuitive than the case above. In Table 1, fire is the only event we know which can cause both smoke and alarm with high certainty ($p(s|f) = p(a|f) = e_2$). Thus observing both simultaneously we would expect a higher probability. But the remaining 8 legal FTOPAs give only ambiguous $p(f|s\&a)$ spanning at least half of the total probability range.

  Consider the evaluation of $p(f|s\&a)$ in model $M_{8,4}$ with idempotent elements $\{e_1, e_4, e_7, e_8\}$.

  $$p(f|s\&a) = p(s|f\&a) * p(f|a)/p(s|a)$$
  $$= e_2 * e_5/e_5 = e_5/e_5 = [e_4, e_1]$$

  Notice the solution in last step.

- In the deductive case, the situation is slightly better. Some models achieve the same tendency as [0, 1] probability in deduction (e.g. $p(s|t) < p(a|t)$). Some achieve the same tendency with increased ambiguity. Others either produce identical ranges for different probabilities or do not reflect the correct trend. The slight improvement attributes to less operations required in deduction (only reasoning by cases but not Bayes theorem is involved). Since reasoning by cases needs the solution operation, it still creates denominator-indifference and generates ambiguity.

Our experiment is systematic with respect to legal FTOPAs of a particular size 8. Although a set of arbitrarily chosen priors is used in this presentation, we have tried varying them in a non-systematic way, but the outcomes were basically the same.

## 6 Conclusion

The investigation is motivated by finding finite totally ordered probability models under the theory of probabilistic logic [Aleliunas 88], to automate qualitative reasoning under uncertainty and facilitate knowledge acquisition and explanation in expert system building.

Under the theory of probabilistic logic, the general form of finite totally ordered probability algebras was derived and the number of different models is deduced such that all the possible models can be explored systematically.

Two major problems of those models are analyzed: denominator-indifference, and ambiguity-generation. They are manifested during the processes of applying Bayes theorem and reasoning by cases. Changes in size, model and assignment of priors do not seem to solve the problems.

All the models with size 8 have been implemented in a Prolog program and tested against a simple example. The results are consistent with the analysis.

The investigation reveals that under the TPL axioms, *finite* probability models may have limited usefulness. The premise of legal FTOPA is {TPL axioms, finite, totally ordered}. It is believed that TPL axioms represent the necessity of general inference under uncertainty. "Totally ordered" seems to be necessary, and is not the real culprit here. Thus it is conjectured that a useful uncertainty management mechanism can not be realized in a finite setting.

## Acknowledgements

This work is supported by Operating Grants A3290 and OGPOO44121 from NSERC. Y. Xiang was awarded a University Fellowship during the term of this work. The authors would like to thank R. Aleliunas for helping us to gain the understanding of his TPL.

## References

[Aleliunas 86] R. Aleliunas, "Models of reasoning based on formal deductive probability theories," *Draft unpublished*, 1986.

[Aleliunas 87] R. Aleliunas, "Mathematical models of reasoning - competence models of reasoning about propositions in English and their relationship to the concept of probability," *Research Report CS-87-31, Univ. of Waterloo*, 1987.

[Aleliunas 88] R. Aleliunas, "A new normative theory of probabilistic logic," *Proc. CSCSI-88*, pp. 67-74, 1988.

[Burris 81] S. Burris and H. P. Sankappannvar, *A course in universal algebra*, Springer-Verlag, 1981.

[Kuczkowski 77] J. E. Kuczkowski and J. L. Gersting, *Abstract Algebra*, Marcel Dekker, 1977.

[Halpern 87] J. Y. Halpern and M.O. Rabin, "A logic to reason about likelihood," *Artificial Intelligence*, 32: 379-405, 1987.

[Pearl 88] J. Pearl *Probabilistic Reasoning in Intelligent Systems: Networks of Plausible Inference*, Morgan Kaufmann.

[Pearl 89] J. Pearl, "Probabilistic semantics for non-monotonic reasoning: A survey," to appear in *Proceedings, First intl. conf. on principles of knowledge representation and reasoning*, 1989.

[Poole 88] D. Poole and E. Neufeld, "Sound probabilistic inference in Prolog: an executable specification of influence diagrams," *I SIMPOSIUM INTERNACIONAL DE INTELIGENCIA ARTIFICIAL*, Oct. 1988.

## Appendix A: Derivation of $p(fire|smoke \& alarm)$

$$p(f|s\&a) = p(s|f\&a) * p(f|a)/p(s|a)$$

where

$$p(s|f\&a) = p(s|f);$$
$$p(s|a) = p(s\&(f \vee \overline{f})|a)$$
$$= i[i[p(s|\overline{f})] *$$
$$i[p(s|f) * p(f|a)/i[p(s|\overline{f}) * p(\overline{f}|a)]]];$$

and

$$p(f|a) = p(a|f) * p(f)/p(a)$$

where

$$p(a|f) = p(a\&((f\&t) \vee (f\&\overline{t}) \vee$$
$$(\overline{f}\&t) \vee (\overline{f}\&\overline{t}))|f)$$
$$= i[i[p(a|f\&\overline{t}) * p(\overline{t})] * i[p(a|f\&t) *$$
$$p(t)/i[p(a|f\&\overline{t}) * p(\overline{t})]]]$$

and

$$p(a) = p(a\&((f\&t) \vee (f\&\overline{t}) \vee$$
$$(\overline{f}\&t) \vee (\overline{f}\&\overline{t})))$$
$$= i[f_1 * f_2 * f_3 * f_4]$$

where

$$f_1 = i[p(a|\overline{f}\&\overline{t}) * p(\overline{f}) * p(\overline{t})]$$
$$f_2 = i[p(a|\overline{f}\&t) * p(\overline{f}) * p(t)/f_1]$$
$$f_3 = i[p(a|f\&\overline{t}) * p(f) * p(\overline{t})/(f_1 * f_2)]$$
$$f_4 = i[p(a|f\&t) * p(f) * p(t)/(f_1 * f_2 * f_3)].$$

# Appendix B: Examples of legal FTOPAs

|       | $e_1$ | $e_2$ | $e_3$ | $e_4$ | $e_5$ | $e_6$ | $e_7$ | $e_8$ |
|-------|-------|-------|-------|-------|-------|-------|-------|-------|
| $e_1$ | $e_1$ | $e_2$ | $e_3$ | $e_4$ | $e_5$ | $e_6$ | $e_7$ | $e_8$ |
| $e_2$ | $e_2$ | $e_3$ | $e_4$ | $e_5$ | $e_6$ | $e_7$ | $e_7$ | $e_8$ |
| $e_3$ | $e_3$ | $e_4$ | $e_5$ | $e_6$ | $e_7$ | $e_7$ | $e_7$ | $e_8$ |
| $e_4$ | $e_4$ | $e_5$ | $e_6$ | $e_7$ | $e_7$ | $e_7$ | $e_7$ | $e_8$ |
| $e_5$ | $e_5$ | $e_6$ | $e_7$ | $e_7$ | $e_7$ | $e_7$ | $e_7$ | $e_8$ |
| $e_6$ | $e_6$ | $e_7$ | $e_7$ | $e_7$ | $e_7$ | $e_7$ | $e_7$ | $e_8$ |
| $e_7$ | $e_7$ | $e_7$ | $e_7$ | $e_7$ | $e_7$ | $e_7$ | $e_7$ | $e_8$ |
| $e_8$ | $e_8$ | $e_8$ | $e_8$ | $e_8$ | $e_8$ | $e_8$ | $e_8$ | $e_8$ |

| $q \backslash p$ | $e_1$ | $e_2$ | $e_3$ | $e_4$ | $e_5$ | $e_6$ | $e_7$ | $e_8$ |
|------|-------|-------|-------|-------|-------|-------|-------|-------|
| $e_1$ | $e_1$ | $e_2$ | $e_3$ | $e_4$ | $e_5$ | $e_6$ | $e_7$ | $e_8$ |
| $e_2$ |       | $e_1$ | $e_2$ | $e_3$ | $e_4$ | $e_5$ | $[e_7,e_6]$ | $e_8$ |
| $e_3$ |       |       | $e_1$ | $e_2$ | $e_3$ | $e_4$ | $[e_7,e_5]$ | $e_8$ |
| $e_4$ |       |       |       | $e_1$ | $e_2$ | $e_3$ | $[e_7,e_4]$ | $e_8$ |
| $e_5$ |       |       |       |       | $e_1$ | $e_2$ | $[e_7,e_3]$ | $e_8$ |
| $e_6$ |       |       |       |       |       | $e_1$ | $[e_7,e_2]$ | $e_8$ |
| $e_7$ |       |       |       |       |       |       | $[e_7,e_1]$ | $e_8$ |

Solution table $p/q$

$M_{8,3}$

|       | $e_1$ | $e_2$ | $e_3$ | $e_4$ | $e_5$ | $e_6$ | $e_7$ | $e_8$ |
|-------|-------|-------|-------|-------|-------|-------|-------|-------|
| $e_1$ | $e_1$ | $e_2$ | $e_3$ | $e_4$ | $e_5$ | $e_6$ | $e_7$ | $e_8$ |
| $e_2$ | $e_2$ | $e_2$ | $e_3$ | $e_4$ | $e_5$ | $e_6$ | $e_7$ | $e_8$ |
| $e_3$ | $e_3$ | $e_3$ | $e_3$ | $e_4$ | $e_5$ | $e_6$ | $e_7$ | $e_8$ |
| $e_4$ | $e_4$ | $e_4$ | $e_4$ | $e_4$ | $e_5$ | $e_6$ | $e_7$ | $e_8$ |
| $e_5$ | $e_5$ | $e_5$ | $e_5$ | $e_5$ | $e_5$ | $e_6$ | $e_7$ | $e_8$ |
| $e_6$ | $e_6$ | $e_6$ | $e_6$ | $e_6$ | $e_6$ | $e_6$ | $e_7$ | $e_8$ |
| $e_7$ | $e_7$ | $e_7$ | $e_7$ | $e_7$ | $e_7$ | $e_7$ | $e_7$ | $e_8$ |
| $e_8$ | $e_8$ | $e_8$ | $e_8$ | $e_8$ | $e_8$ | $e_8$ | $e_8$ | $e_8$ |

| $q \backslash p$ | $e_1$ | $e_2$ | $e_3$ | $e_4$ | $e_5$ | $e_6$ | $e_7$ | $e_8$ |
|------|-------|-------|-------|-------|-------|-------|-------|-------|
| $e_1$ | $e_1$ | $e_2$ | $e_3$ | $e_4$ | $e_5$ | $e_6$ | $e_7$ | $e_8$ |
| $e_2$ |       | $[e_2,e_1]$ | $e_3$ | $e_4$ | $e_5$ | $e_6$ | $e_7$ | $e_8$ |
| $e_3$ |       |       | $[e_3,e_1]$ | $e_4$ | $e_5$ | $e_6$ | $e_7$ | $e_8$ |
| $e_4$ |       |       |       | $[e_4,e_1]$ | $e_5$ | $e_6$ | $e_7$ | $e_8$ |
| $e_5$ |       |       |       |       | $[e_5,e_1]$ | $e_6$ | $e_7$ | $e_8$ |
| $e_6$ |       |       |       |       |       | $[e_6,e_1]$ | $e_7$ | $e_8$ |
| $e_7$ |       |       |       |       |       |       | $[e_7,e_1]$ | $e_8$ |

Solution table $p/q$

$M_{8,8}$

|       | $e_1$ | $e_2$ | $e_3$ | $e_4$ | $e_5$ | $e_6$ | $e_7$ | $e_8$ |
|-------|-------|-------|-------|-------|-------|-------|-------|-------|
| $e_1$ | $e_1$ | $e_2$ | $e_3$ | $e_4$ | $e_5$ | $e_6$ | $e_7$ | $e_8$ |
| $e_2$ | $e_2$ | $e_3$ | $e_4$ | $e_4$ | $e_5$ | $e_6$ | $e_7$ | $e_8$ |
| $e_3$ | $e_3$ | $e_4$ | $e_4$ | $e_4$ | $e_5$ | $e_6$ | $e_7$ | $e_8$ |
| $e_4$ | $e_4$ | $e_4$ | $e_4$ | $e_4$ | $e_5$ | $e_6$ | $e_7$ | $e_8$ |
| $e_5$ | $e_5$ | $e_5$ | $e_5$ | $e_5$ | $e_6$ | $e_7$ | $e_7$ | $e_8$ |
| $e_6$ | $e_6$ | $e_6$ | $e_6$ | $e_6$ | $e_7$ | $e_7$ | $e_7$ | $e_8$ |
| $e_7$ | $e_7$ | $e_7$ | $e_7$ | $e_7$ | $e_7$ | $e_7$ | $e_7$ | $e_8$ |
| $e_8$ | $e_8$ | $e_8$ | $e_8$ | $e_8$ | $e_8$ | $e_8$ | $e_8$ | $e_8$ |

| $q \backslash p$ | $e_1$ | $e_2$ | $e_3$ | $e_4$ | $e_5$ | $e_6$ | $e_7$ | $e_8$ |
|------|-------|-------|-------|-------|-------|-------|-------|-------|
| $e_1$ | $e_1$ | $e_2$ | $e_3$ | $e_4$ | $e_5$ | $e_6$ | $e_7$ | $e_8$ |
| $e_2$ |       | $e_1$ | $e_2$ | $[e_4,e_3]$ | $e_5$ | $e_6$ | $e_7$ | $e_8$ |
| $e_3$ |       |       | $e_1$ | $[e_4,e_2]$ | $e_5$ | $e_6$ | $e_7$ | $e_8$ |
| $e_4$ |       |       |       | $[e_4,e_1]$ | $e_5$ | $e_6$ | $e_7$ | $e_8$ |
| $e_5$ |       |       |       |       | $[e_4,e_1]$ | $e_5$ | $[e_7,e_6]$ | $e_8$ |
| $e_6$ |       |       |       |       |       | $[e_4,e_1]$ | $[e_7,e_5]$ | $e_8$ |
| $e_7$ |       |       |       |       |       |       | $[e_7,e_1]$ | $e_8$ |

Solution table $p/q$

One of $M_{8,4}$ with idempotent elements $e_1, e_4, e_7$ and $e_8$